%% file: root.tex
\documentclass[letterpaper, 10 pt, conference]{ieeeconf}  

\IEEEoverridecommandlockouts                              

\usepackage{amsmath} 
\usepackage{amssymb}  

\usepackage{amsthm}

\usepackage{cite}
\usepackage{dsfont}
\usepackage{graphicx}
\usepackage[ruled,vlined]{algorithm2e}
\usepackage[bookmarks=false]{hyperref}
\usepackage{xcolor}
\usepackage{booktabs}
\usepackage{colortbl}
\definecolor{drpablue}{RGB}{0, 102, 204}
\newcommand{\PP}[1]{\left(#1\right)}

\newtheorem*{property1}{Property 1}
\newtheorem*{property2}{Property 2}
\usepackage{threeparttable}

\makeatletter
\let\MYcaption\@makecaption
\makeatother

\usepackage[font=footnotesize]{subcaption}

\makeatletter
\let\@makecaption\MYcaption
\makeatother

\usepackage{subcaption}

\title{\LARGE \bf
DRPA-MPPI: Dynamic Repulsive Potential Augmented MPPI \\ for Reactive Navigation in Unstructured Environments
}

\author{Takahiro Fuke$^{1}$,  Masafumi Endo$^{1}$, Kohei Honda$^{2}$, and Genya Ishigami$^{1}$
\thanks{*This work was not supported by any organization}
\thanks{$^{1}$T. Fuke, M. Endo, and G. Ishigami are with the Space Robotics Group, Department of Mechanical Engineering, Keio University, Yokohama 223-8522, Japan
        {\tt\small 
        \{taka162uke, masafumi.endo\}@keio.jp,
        ishigami@mech.keio.ac.jp}
        }
\thanks{$^{2}$K. Honda is with the Mobility System Group, Department of Mechanical Systems Engineering, Nagoya University, Nagoya 464-8603, Japan 
        {
        \tt\small
        honda.kohei.k3@f.mail.nagoya-u.ac.jp
        }
        }%
}

\begin{document}

\twocolumn[
\noindent
© 2025 IEEE. Personal use of this material is permitted. Permission from IEEE must be obtained for all other uses, in any current or future media, including reprinting/republishing this material for advertising or promotional purposes, creating new collective works, for resale or redistribution to servers or lists, or reuse of any copyrighted component of this work in other works.\\

\noindent
\textbf{Submitted article:}\\
T. Fuke, M. Endo, K. Honda, and G. Ishigami ``DRPA-MPPI: Dynamic Repulsive Potential Augmented MPPI for Reactive Navigation in Unstructured Environments
,'' \textit{IEEE International Conference on Automation Science and Engineering}, 2025.
]
\thispagestyle{empty}
\pagenumbering{gobble}
\clearpage

\maketitle
\thispagestyle{empty}
\pagestyle{empty}


\begin{abstract}
Reactive mobile robot navigation in unstructured environments is challenging when robots encounter unexpected obstacles that invalidate previously planned trajectories. 
Model predictive path integral control (MPPI) enables reactive planning, but still suffers from limited prediction horizons that lead to local minima traps near obstacles.
Current solutions rely on heuristic cost design or scenario-specific pre-training, which often limits their adaptability to new environments. 
We introduce dynamic repulsive potential augmented MPPI (DRPA-MPPI), which dynamically detects potential entrapments on the predicted trajectories.
Upon detecting local minima, DRPA-MPPI automatically switches between standard goal-oriented optimization and a modified cost function that generates repulsive forces away from local minima. 
Comprehensive testing in simulated obstacle-rich environments confirms DRPA-MPPI's superior navigation performance and safety compared to conventional methods with less computational burden.
\end{abstract}


\input{src/sec1_introduction}
\input{src/sec2_related_works}
\input{src/sec3_preliminaries}
\input{src/sec4_drpa_mppi}
\input{src/sec5_experiments}
\input{src/sec6_conclusion} 

\addtolength{\textheight}{-0cm}   


\input{src/appendix}


\bibliographystyle{IEEEtran}
\bibliography{IEEEexample}
\end{document}

%% file: src/sec1_introduction.tex
\section{Introduction}\label{sec1}

\looseness=-1
Modern robot navigation heavily relies on the availability of sufficient environmental information.
However, in unstructured environments, such information is often incomplete or unreliable. 
For example, in robotic exploration or rescue missions under unknown scenarios, where sparse (sub-)goals are provided as coordinates, but the obstacles and layout between these points may be inaccessible or outdated. 
In such scenarios, the robot must navigate to its goal while detouring around obstacles, even though global planners cannot offer collision-free paths due to the lack of detailed environmental information. 
Therefore, a standalone reactive local planner is essential to enable the robot to avoid obstacles and reach its destination without relying on pre-planned global paths.

\begin{figure}
    \centering
    \includegraphics[width=\linewidth]{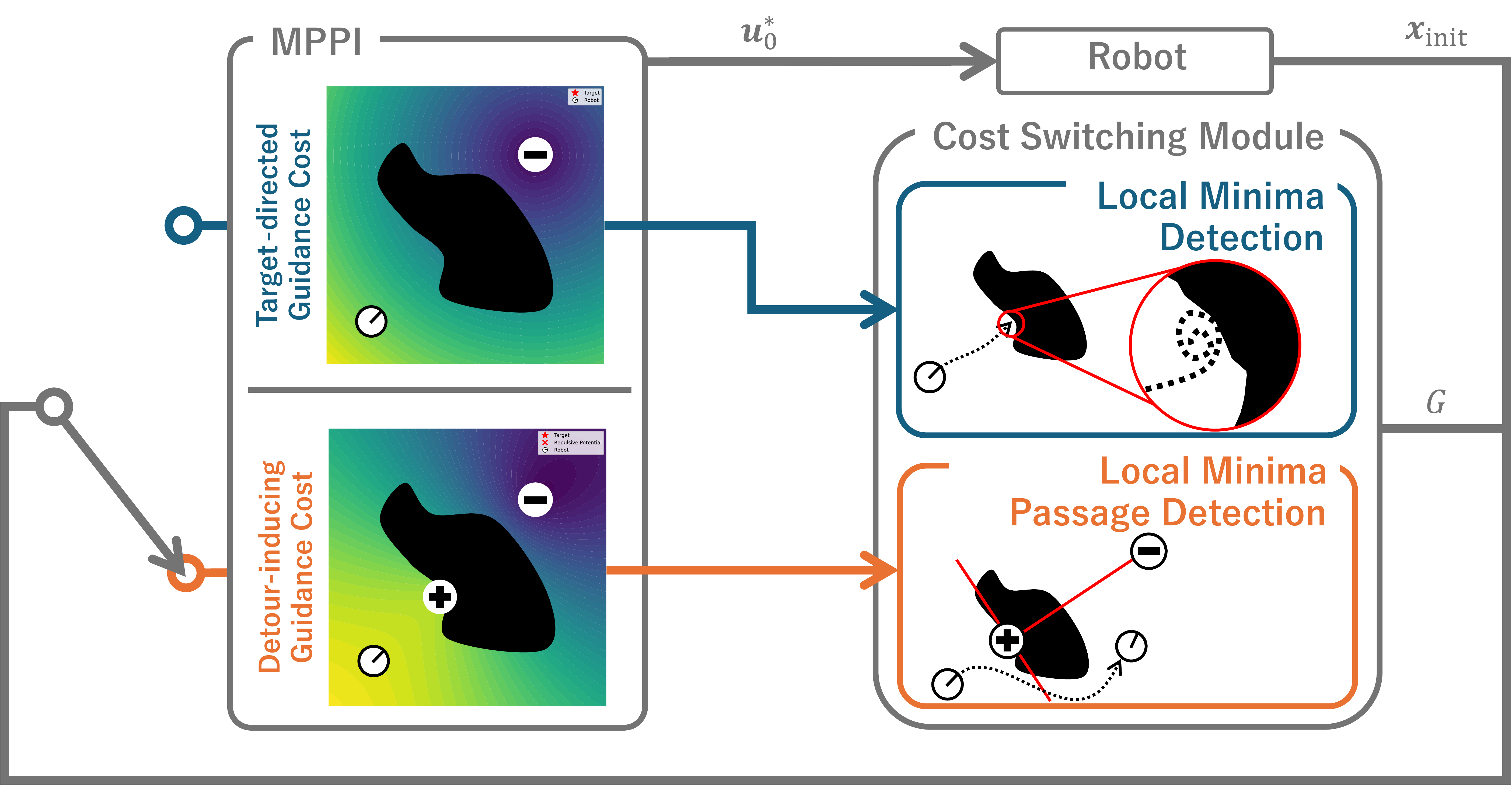}
    \caption{Overview of the DRPA-MPPI framework. The framework switches between target-directed and detour-inducing guidance costs based on local minima detection and local minima passage detection.}
    \label{sec1:fig:overview}
\end{figure}

A common method for reactive local planning is sampling-based model predictive control (MPC)~\cite{sampling_based_mpc_survey}, such as model predictive path integral control (MPPI)~\cite{mppi}, which minimizes a finite-horizon cost function through a stochastic, sample-based search performed in real-time. 
While MPPI can account for unexpected obstacles by incorporating non-differentiable sensor data into its cost function, it remains susceptible to local minima—suboptimal positions that appear optimal within a limited area—especially near large or non-convex obstacles. 
To address this limitation, previous studies have explored both rule-based~\cite{subgoal_sgp_mppi, cost_search_based_mpc, cost_sample_based_mppi, sampling_dist_rrt_mppi} and data-driven~\cite{subgoal_gonext_mpc, cost_polo_mpc, cost_mpq_mppi, sampling_dist_flow_mppi} approaches. 
However, these methods often require the heuristic design of scenario-specific controllers or pre-training. 
Similarly, our previous work~\cite{rpa_mppi} proposed a method for escaping local minima, but relied on heuristics that assumed prior knowledge of local minima positions.

To this end, we propose an automatic local minima detection and detour method for reactive local planning without scenario-specific designs, called dynamic repulsive potential augmented MPPI (DRPA-MPPI), as shown in Fig.~\ref{sec1:fig:overview}.
DRPA-MPPI builds upon the concept of our previous work~\cite{rpa_mppi} by adding local minima detection for reactive navigation in unstructured environments with unpredictable local minima.
Our key ideas shown in Fig. \ref{sec1:fig:overview} include: 1) \emph{dynamic} local minima detection that analyzes predicted robot trajectories to identify entrapment situations; 2) \emph{repulsive potential augmentation} that adds terms to the cost function to generate detouring behaviors from detected local minima; and 3) integrating these components into MPPI while maintaining computational efficiency.
Upon detecting potential entrapment, the framework switches from a target-directed cost function to one with repulsion from the local minimum.

We evaluate our method through simulated environments with various obstacles, including non-convex ones, which introduce potential local minima entrapment.
Experimental results demonstrate that our method outperforms existing methods in terms of navigation safety and efficiency while preserving computational efficiency.
Theoretical analysis also guarantees its ability to avoid entrapment in local minima. 

%% file: src/sec2_related_works.tex
\section{Related Work}\label{sec2}

In this work, we tackle navigation without a global path, where only the goal position is provided and no prior map exists between the current position and goal.
This task requires reactive local planning to steer toward the goal while avoiding unknown obstacles.
The following section first describes the background and local minima issues of MPC-based reactive planning, then positions our work among existing approaches to local minima avoidance.

\subsection{MPC-based Reactive Local Planning}

Reactive local planning generates control commands based primarily on onboard sensor information to avoid obstacles and guide the robot to the goal. 
MPC is a powerful tool for such planning as it incorporates predictions of the robot's future states.
Sampling-based MPCs~\cite{sampling_based_mpc_survey, dwa}, rather than gradient-based ones~\cite{mpc_survey_1,mpc_survey_2,mpc_survey_3}, have proven more practical for robotic navigation because they directly handle sensor information as cost maps~\cite{macenski2023desks}. 
Among these, MPPI~\cite{mppi} is promising, using stochastic sampling and GPU parallelization to efficiently explore solution spaces. 
However, when a non-convex obstacle lies between the robot and the goal, standard MPPI local planners are entrapped in local minima. 
The challenge arises from the difficulty of finding low-cost detours around obstacles within the finite prediction horizon.

\subsection{Local Minima Avoidance in MPC-based Local Planners}

\looseness=-1
To address the aforementioned local minima issue, two main approaches have been developed: sub-goal recommendation and cost-to-go heuristics.
The former approach introduces intermediate waypoints between the robot's current position and the final goal.
Mohamed \emph{et al.} proposed a local perceptual model using sparse Gaussian processes for the sub-goal recommendation~\cite{subgoal_sgp_mppi}.
Brito \emph{et al.} developed a data-driven method using reinforcement learning while it struggles to generalize beyond its training scenarios~\cite{subgoal_gonext_mpc}.
These external modules introduce additional complexity in the planning framework and highlight the need to improve the inherent robustness of local planners against local minima, rather than relying on auxiliary systems.

Alternatively, cost-to-go heuristics modify the planning objective to better approximate globally-optimal behavior. 
These heuristics fall into two categories: data-driven approaches using reinforcement learning for value estimation~\cite{cost_polo_mpc, cost_mpq_mppi}, and rule-based approaches integrating global planners~\cite{cost_search_based_mpc, cost_sample_based_mppi}.
Additionally, some studies focus on adaptive modifications to MPPI sampling distributions, such as those that use generative models~\cite{sampling_dist_flow_mppi} or employ global planners~\cite{sampling_dist_rrt_mppi}.
However, data-driven methods require extensive pretraining and often fail in new environments, and methods using global planners perform poorly in unstructured environments where prior map information is unreliable.

We propose a practical solution with an event-triggered switching module in the MPPI framework. 
This module detects potential local minima entrapments and automatically switches between two distinct cost functions.
Our approach is simple yet effective, unlike previous work that relies on data-driven methods~\cite{cost_polo_mpc, cost_mpq_mppi} or integrates with global planners~\cite{cost_search_based_mpc, cost_sample_based_mppi}.
It improves the resilience of MPPI to local minima, while maintaining competitive obstacle avoidance performance and computational efficiency.

%% file: src/sec3_preliminaries.tex
\section{Standard MPPI-based Local Planning}\label{sec3}
{

    This section presents the fundamental principles of MPPI and its standard cost formulation for local planning.
    \subsection{Review of MPPI}\label{sec3:sec:fundamentals}
    {
        MPPI considers nonlinear dynamical systems with Gaussian noise added to the control input:
        \begin{equation}\label{sec3:eq:stochastic_dynamics}
            {\bf{x}}_{\tau+1} = {\bf{F}}({\bf{x}}_{\tau},{\bf{u}}_{\tau}+\delta{\bf{u}}_{\tau}),
        \end{equation}
        where ${\bf{x}}_{\tau}\in\mathbb{R}^{n}$, ${\bf{u}}_{\tau}\in\mathbb{R}^{m}$, and $\delta{\bf{u}}_{\tau}\in\mathbb{R}^{m}$ represent the state, control input, and Gaussian control noise with zero mean and covariance matrix $\boldsymbol{\Sigma}$, respectively. 
        For a finite horizon $\tau\in\{0,1,\cdots,T\}$, the stochastic optimal control problem seeks to find the optimal control sequence ${\mathbf{U}}^{*}=\{{\bf{u}}_{0}^{*},{\bf{u}}_{1}^{*},\cdots,{\bf{u}}_{T-1}^{*}\}$ by solving:
        \begin{subequations}
        \begin{equation}
            {\mathbf{U}}^{*}=\underset{\mathbf{U}}{\operatorname{argmin}}\,
            \mathbb{E}\left[\phi({\bf{x}}_{T}) + \sum_{\tau = 0}^{T-1}\left(c({\bf{x}}_{\tau}) + \frac{\lambda}{2}{\bf{u}}_{\tau}^{\rm{T}}\boldsymbol{\Sigma}^{-1}{\bf{u}}_{\tau}\right)\right], \\
            \label{sec3:eq:optimization_problem}
        \end{equation}
        subject to
        \begin{align}
            & {\bf{x}}_{\tau+1} = {\bf{F}}({\bf{x}}_{\tau}, {\bf{u}}_\tau + \delta{\bf{u}}_\tau),\, \delta {\bf{u}}_{\tau}\sim \mathcal{N}(0,\boldsymbol{\Sigma}), \label{sec3:eq:dynamics_constraint}& \\
            & {\bf{x}}_{0} = {\bf{x}}_{\rm{init}}. & \label{sec3:eq:initial_condition}
        \end{align}
        \end{subequations}
        Here, $\phi(\mathbf{x}_{T})$ and $c(\mathbf{x}_{\tau})$ represent the arbitrary terminal and state-dependent running costs, respectively.
        The term following the state-dependent running cost is the control cost, expressed as a quadratic form of the control input. 
        ${\bf{x}}_{\rm{init}}$ denotes the initial state of the system.
        
        MPPI solves the optimization problem in \eqref{sec3:eq:optimization_problem} through an importance sampling technique~\cite{kloek1978bayesian}.
        The method begins with an initial estimated control sequence ${\mathbf{\widehat{U}}}=\{\widehat{{\bf{u}}}_{0},\widehat{{\bf{u}}}_{1},\cdots,\widehat{{\bf{u}}}_{T-1}\}$. 
        It then generates $K$ perturbed control sequences $\mathbf{U}_k$ by adding Gaussian noise $\boldsymbol{\epsilon}\sim\mathcal{N}(0,\boldsymbol{\Sigma}_{\boldsymbol{\epsilon}})$ to each components of $\mathbf{\widehat{U}}$, where $\boldsymbol{\Sigma}_{\boldsymbol{\epsilon}}$ is the covariance matrix for the sampling distribution. 
        These perturbed sequences simulate $K$ system trajectories, or \emph{rollouts}. 
        The costs associated with these rollouts are used to compute a weighted average, which forms the basis for updating the control sequence.
        The cost-to-go of the $k$-th rollout is computed as:
        \begin{equation}
        J (\mathbf{x}_\text{init}, \mathbf{U}_k) = \phi({\bf{x}}_{k,T}) + \sum_{\tau = 0}^{T-1}\left(c({\bf{x}}_{k,\tau}) + {\gamma}{\widehat{\bf{u}}_{\tau}}^{\rm{T}}\boldsymbol{\Sigma}_{\epsilon}^{-1}{\bf{u}}_{k,\tau}\right), \label{sec3:eq:rollout_cost}
        \end{equation}
        and the optimal control input ${\bf{u}}_{\tau}^{*}$ is updated as follows:
        \begin{equation}\label{sec3:eq:control_update}
            {\bf{u}}_{\tau}^{*}={\widehat{\bf{u}}_{\tau}}+\frac{\sum_{k = 0}^{K-1} \exp{\left(-(1/\lambda) (J (\mathbf{x}_\text{init}, \mathbf{U}_k)-\rho)\right)}\boldsymbol{\epsilon}_{k,\tau}}{\sum_{k = 0}^{K-1} \exp{\left(-(1/\lambda) (J (\mathbf{x}_\text{init}, \mathbf{U}_k)-\rho)\right)}},
        \end{equation}
        where $\lambda \in {\mathbb{R}}^{+}$ is a temperature parameter, $\gamma \in [0,\lambda]$ is a control cost parameter, and $\rho$ is the minimum cost-to-go to prevent numerical instability. 
        The updated solution in \eqref{sec3:eq:control_update} minimizes the Kullback-Leibler divergence with the optimal control distribution characterized by the Boltzmann distribution. 
        The rollouts in MPPI are independent of one another, which allows efficient GPU-based parallel computation. 
        MPPI also does not rely on gradient calculations and iterative solution updates. 
        This flexibility enables the use of arbitrary prediction models and cost designs, making MPPI well-suited for navigation in unstructured environments.
    }
        
    \subsection{Standard Cost Design for Reactive Local Planning}\label{sec3:sec:target_cost}
    {
        A basic structure for navigation tasks includes a position-based state-dependent running cost $c(\mathbf{x}_{\tau})$ and terminal cost $\phi(\mathbf{x}_{T})$ defined as:
        \begin{subequations}
        \begin{align}
            &c(\mathbf{x}_{\tau})
            = w_{\text{obst}} \cdot \mathds{1}_{\text{obst}}(\mathbf{p}_{\tau}), \\
            &\phi(\mathbf{x}_{T}) = w_{\text{obst}} \cdot \mathds{1}_{\text{obst}}(\mathbf{p}_{T})+w_{\text{guidance}}\cdot G(\mathbf{p}_{T}).
            \label{sec3:eq:general_cost}
        \end{align}
        \end{subequations}
        Here, $\mathbf{p}$ represents position component of the state $\mathbf{x}$. 
        Although non-positional state variables could improve control stability, we focus on position-based terms for simplicity.

        The first term in both equations represents the collision cost, with $w_{\text{obst}} \in {\mathbb{R}}^{+}$ as a collision weight and $\mathds{1}_{\text{obst}}$ as an indicator function defined using the set of positions $\mathbb{O}_{\text{obst}}$ occupied by obstacles as:
        
        \begin{equation}\label{sec3:eq:indicator_function}
            \mathds{1}_{\text{obst}}(\mathbf{p}) = \begin{cases}
                1 & \text{if } \mathbf{p} \in \mathbb{O}_{\text{obst}} \\
                0 & \text{if } \mathbf{p} \notin \mathbb{O}_{\text{obst}}
            \end{cases}.
        \end{equation}
    
        The second term in the terminal cost represents the guidance cost, where $w_{\text{guidance}} \in {\mathbb{R}}^{+}$ is the guidance weight and $G$ is the guidance function. 
        Following convention, this function uses the Euclidean distance to target position $\mathbf{p}_{\text{target}}$:
        \begin{equation}\label{sec3:eq:target_directed_function}
            G_{\text{target}}(\mathbf{p}) = \|\mathbf{p}_{\text{target}} - \mathbf{p}\|.
        \end{equation}
        
        The target-directed guidance function $G_{\text{target}}$ formulated as the Euclidean distance in \eqref{sec3:eq:target_directed_function} is effective and widely used for navigation around relatively small or convex obstacles. 
        However, it shows limitations when dealing with larger or non-convex obstacles. 
        In such scenarios, the robot becomes trapped in local minima, unable to execute necessary detours to reach the target position. 
        We are motivated to overcome local minima problems without the limitations of previous methods, which used subgoal recommendations or heuristic improvements but faced drawbacks discussed in Section \ref{sec1}.
    }

%% file: src/sec4_drpa_mppi.tex
\section{DRPA-MPPI}\label{sec4}
{
    Fig.~\ref{sec1:fig:overview} illustrates the proposed DRPA-MPPI framework. 
    Our approach extends MPPI-based navigation by introducing a mode-switching module. 
    The framework primarily operates with the target-directed guidance function defined in \eqref{sec3:eq:target_directed_function} while continuously monitoring the predicted trajectory to detect potential entrapment in local minima. 
    Upon detecting such entrapment, a detour-inducing guidance function is activated to generate repulsion from the detected local minimum. 
    After successfully passing through the problematic region, the framework reverts to the target-directed guidance function. 
    This dynamic switching continues until the robot reaches its target. 
    In the following subsections, we detail the three core components of our framework: local minimum detection (Sec. \ref{sec4_subsec:local minima detection}), detour-inducing guidance function (Sec. \ref{sec4_subsec:detour-inducing function}), and local minimum passage detection (Sec. \ref{sec4_subsec:termination}). The complete algorithmic implementation integrating these components is also presented (Sec.~\ref{sec4_subsec:drpa-framework}).
    
    \subsection{Local Minima Detection}\label{sec4_subsec:local minima detection}
    {
        Our method for detecting potential entrapment in local minima relies on a simple yet effective observation: when trapped, a robot tends to remain confined within a small spatial region. 
        We implement this insight by monitoring the terminal portion of the predicted trajectory.
        
        Let $\tau_{\text{monitor}}\in[0,T]$ denote the starting index of our monitoring window. 
        We focus on the positions from $\tau_{\text{monitor}}$ to $T$ in the predicted trajectory, $\mathbf{P}^{*}_{\text{monitor}} = \{\mathbf{p}_{\tau_{\text{monitor}}}^{*}, \ldots, \mathbf{p}_{T}^{*}\}.$
        We define a variation measure,
        \begin{equation}
          f(\mathbf{P}^{*}_{\text{monitor}})
          = 
          \frac{1}{T - \tau_{\text{monitor}} + 1}
          \sum_{\tau =\tau_{\text{monitor}}}^{T}
          \bigl\|\mathbf{p}_{\tau_{\text{monitor}}}^{*} - \mathbf{p}_{\tau}^{*}\bigr\|,
        \end{equation}
        which quantifies how far the trajectory moves away from the position at $\tau_{\text{monitor}}$. 
        When $f(\mathbf{P}^{*}_{\text{monitor}})$ falls below a threshold $r_{\text{thres}}$, we consider the terminal portion of the predicted trajectory to be trapped in a local minimum. 
        The position of this local minimum is then estimated as the average of all positions within the monitoring window:
        \begin{equation}
          \mathbf{p}_{\text{min}}
          =
          \frac{1}{T - \tau_{\text{monitor}} + 1}
          \sum_{\tau =\tau_{\text{monitor}}}^{T}
          \mathbf{p}_{\tau}^{*}.
        \end{equation}
    }

    \subsection{Detour-Inducing Guidance Function}\label{sec4_subsec:detour-inducing function}
    {
        When potential entrapment in local minimum is detected, the framework switches from the target-directed guidance function to the detour-inducing guidance function. 
        While the former aims to find the shortest path to the target position, the latter is designed to navigate around obstacles that cause local minima entrapment. 
        This detour-inducing guidance function is formulated as:
        
        \begin{equation}
            G_{\text{detour}}(\mathbf{p}) = 
            \|\mathbf{p}_{\text{vt}} - \mathbf{p}\| 
            - w_{\text{rep}}\|\mathbf{p}_{\text{min}} - \mathbf{p}\|,
        \end{equation}
        where $\mathbf{p}_{\text{vt}}$ is the temporary virtual target position, and $w_{\text{rep}}\in(0,1)$ is the weighting coefficient that balances attraction toward the virtual target and repulsion from the detected local minimum. 
        To ensure effective obstacle avoidance, the virtual target is positioned at a fixed distance $d_{\text{vt}}$ from the detected local minimum toward the original target position:
        
        \begin{equation}
            \mathbf{p}_{\text{vt}} = \mathbf{p}_{\text{min}} + \frac{\mathbf{p}_{\text{target}} - \mathbf{p}_{\text{min}}}{\|\mathbf{p}_{\text{target}} - \mathbf{p}_{\text{min}}\|} d_{\text{vt}}.
        \end{equation}
        
        This detour-inducing guidance function exhibits two theoretical properties:
        \begin{property1}
        {
            \(\mathbf{p}_{\text{vt}}\) is both the unique local and global minimum when \(0 < w_{\text{rep}} < 1\).
        }
        \end{property1}    
        \begin{property2}
        {
            Suppose $\mathbf{p}_{\text{n}}$ satisfies $(\mathbf{p}_{\textup{vt}} - \mathbf{p}_{\textup{min}}) \cdot (\mathbf{p}_{\text{n}} - \mathbf{p}_{\textup{min}}) = 0$ and $\mathbf{p}_{\text{n}} \neq \mathbf{p}_{\text{min}}$. 
            Then $-\nabla G_{\textup{detour}}(\mathbf{p}_{\text{n}}) \cdot (\mathbf{p}_{\text{n}} - \mathbf{p}_{\textup{min}}) \geq 0$ holds if $\|\mathbf{p}_{\textup{min}} - \mathbf{p}_{\text{n}}\| \leq d_{\textup{virtual}} \PP{\frac{w_{\textup{rep}}^2}{1 - w_{\textup{rep}}^2}}^{\frac{1}{2}}$.
        }
        \end{property2}
        Mathematical proofs for both properties are provided in the Appendix. 
        These properties generalize concepts from our previous work~\cite{rpa_mppi} with vector representations and guarantee that with appropriate settings of $w_\text{rep}$ and $d_\text{vt}$, the robot can navigate around any convex obstacle while it maintains progress toward the goal position, as illustrated in Fig. \ref{sec4:fig:compare}. 
        For non-convex obstacles, our approach remains effective in practice, although the same rigorous mathematical guarantees do not apply.
        The combined attractive and repulsive forces guide the robot away from detected local minima and enable successful navigation around complex geometries such as U-shaped obstacles, as shown in Sec. \ref{sec5}.
        
        We simplify the tuning process by setting $w_{\text{rep}} = 0.7$, which gives $\PP{\frac{w_{\text{rep}}^2}{1 - w_{\text{rep}}^2}}^{\frac{1}{2}} \approx 1$. This makes $d_{\text{vt}}$ the primary parameter for adjusting detour behavior, with the boundary of the repulsive region being set at approximately $d_{\text{vt}}$ distance from the local minimum.
    }

    \begin{figure}[t]
        \centering
        \includegraphics[width=\linewidth]{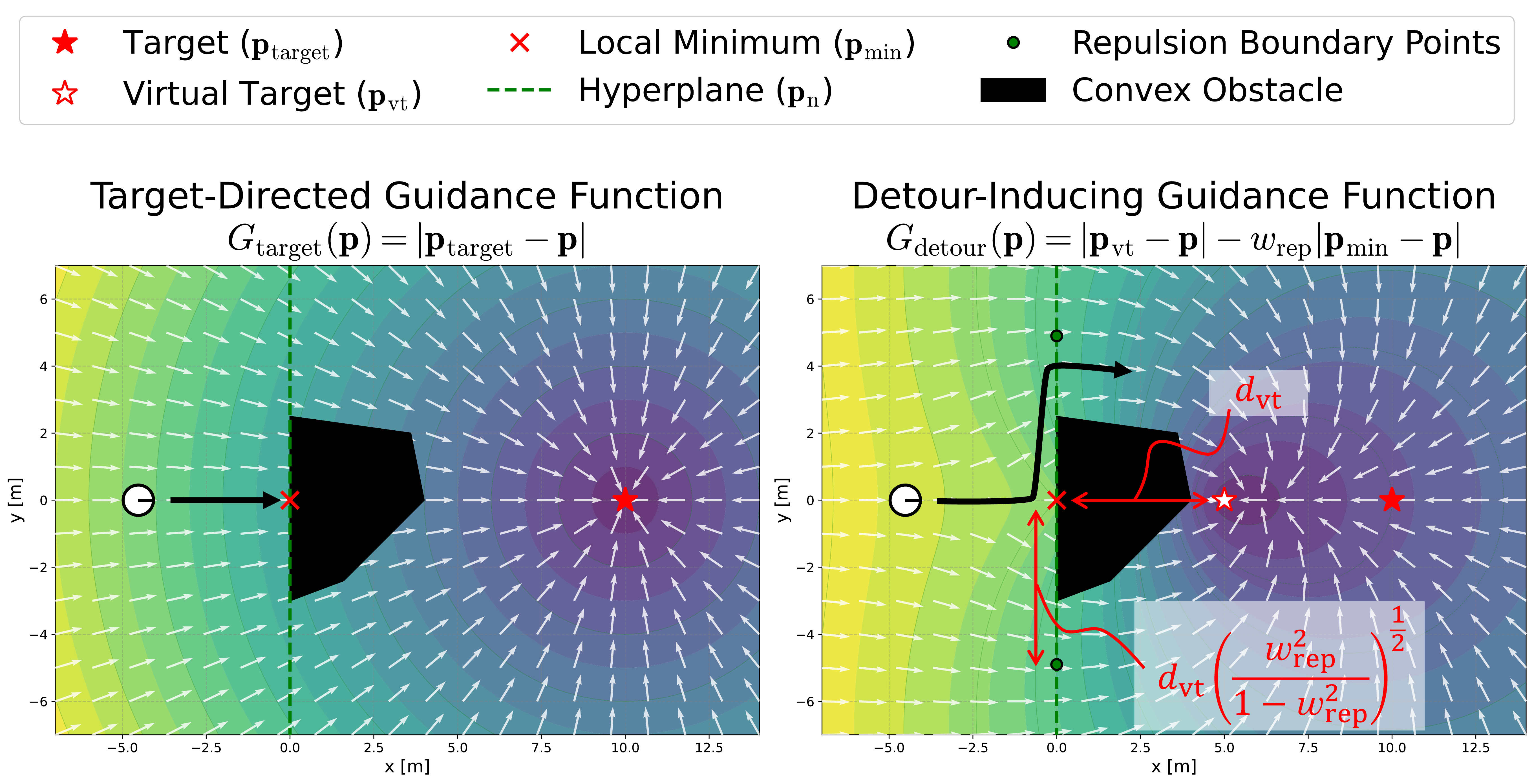}
        \caption{Comparison of guidance functions with a convex obstacle between robot and target. Left: The target-directed guidance function lacks detour gradients along the obstacle edge, which can trap the robot in local minima. Right: The proposed detour-inducing guidance function combines repulsive forces from detected local minima with attractive forces toward a virtual target to create effective obstacle circumnavigation gradients.}
        \label{sec4:fig:compare}
    \end{figure}
    
    \subsection{Local Minima Passage Detection}
    \label{sec4_subsec:termination}
    {
        Our local minimum passage detection uses a geometric criterion.
        Let $\mathbf{p}_{\text{current}}$ denote the robot's current position. 
        The framework monitors the geometric relationship between three key points: the current position, the original target position, and the detected local minimum.
        We define a transition boundary using a hyperplane perpendicular to the vector from the detected local minimum to the target.
        To avoid the mode chattering, this hyperplane is offset from the local minimum by a margin distance $d_{\text{margin}}$ toward the target:
        
        \begin{equation}\label{sec4:eq:margin}
          \mathbf{d}_{\text{margin}} =
          \frac{\mathbf{p}_{\text{target}} - \mathbf{p}_{\text{min}}}{
                \|\mathbf{p}_{\text{target}} - \mathbf{p}_{\text{min}}\|}
          \, d_{\text{margin}}.
        \end{equation}
        If the robot crosses this boundary, the framework will revert to the target-directed cost:
        \begin{equation}\label{sec4:eq:termination}
           (\mathbf{p}_{\text{target}} - \mathbf{p}_{\text{current}})
           \,\cdot\,
           \Bigl(
                 \bigl(\mathbf{p}_{\text{min}} + \mathbf{d}_{\text{margin}}\bigr)- \mathbf{p}_{\text{current}}
           \Bigr) < 0.
        \end{equation}
        This allows the robot to know that it has moved far from the local minima before resuming target-oriented navigation.
        It prevents premature switching that could lead to re-entrapment.
    }

    \begin{algorithm}[t]
        \footnotesize
        \LinesNumbered
        \SetKwInOut{Input}{Given}
        \Input{
            $\mathbf{F}\PP{\cdot, \cdot}$, $c\PP{\cdot}$, $\phi\PP{\cdot}$, $G_{\text{target}}$, $\mathbf{p}_{\text{target}}$, $\hat{\mathbf{U}}$, $K$, $T$, $\lambda$, $\gamma$, $\boldsymbol{\Sigma}_{\epsilon}$,
            \textcolor{drpablue}{$G_{\text{detour}}\PP{\cdot}$, 
            $\tau_{\text{monitor}}$, $r_{\text{thrs}}$, $d_{\text{vt}}$, $d_{\text{margin}}$}
        }
        \BlankLine
        $SetTarget\PP{G_{\text{target}}, \mathbf{p}_{\text{target}}}$\;
        $SetGuidanceCost\PP{\phi,G_{\text{target}}}$\;
        \BlankLine
        \While{robot not reaching target position}{
            $\mathbf{x}_0\leftarrow EstimateState()$\;
            \BlankLine
            $\mathbb{O}_{\text{obst}}\leftarrow EstimateObstacleMap()$\;
            $SetObstacleMap(c,\phi,\mathbb{O}_{\text{obst}})$\;
            \BlankLine
            \tcp{Rollout sampling and cost computation}
            \For{$k \leftarrow 1$ \KwTo $K$ in parallel}{
                $J_k \leftarrow 0$\;
                $\mathbf{x}_{k,0} \leftarrow \mathbf{x}_0$\;
                \For{$\tau \leftarrow 0$ \KwTo $T-1$}{
                    $\boldsymbol{\epsilon}_{k,\tau} \sim \mathcal{N}\PP{\mathbf{0}, \boldsymbol{\Sigma}_{\epsilon}}$\;
                    $\mathbf{v}_{k,\tau} \leftarrow \hat{\mathbf{u}}_{\tau} + \boldsymbol{\epsilon}_{k,\tau}$\;
                    $\mathbf{x}_{k,\tau+1} \leftarrow \mathbf{F}\left(\mathbf{x}_{k,\tau}, \mathbf{v}_{k,\tau}\right)$\;
                    \uIf{$\tau=T-1$}{$J_k \mathrel{+}= \phi\PP{\mathbf{x}_{k,\tau+1}}$\;}
                    \Else{$J_k \mathrel{+}= c\PP{\mathbf{x}_{k,\tau+1}}$\;}
                    \BlankLine
                    $J_k \mathrel{+}= \gamma \hat{\mathbf{u}}_{\tau}^{\mathrm{T}}\boldsymbol{\Sigma}_{\epsilon}^{-1}\mathbf{v}_{k,\tau}$\;
                }
            }
            \BlankLine
            \tcp{Importance sampling and control update}
            $\rho \leftarrow \min\{J_1, \ldots, J_K\}$\;
            $\eta \leftarrow \sum_{k=1}^{K}\exp\Bigl(-\frac{1}{\lambda}(J_k - \rho)\Bigr)$\;
            \For{$k \leftarrow 1$ \KwTo $K$ in parallel}{
                $w_k \leftarrow \frac{1}{\eta} \exp\Bigl(-\frac{1}{\lambda}(J_k - \rho)\Bigr)$\;
            }
            \BlankLine
            \For{$\tau \leftarrow 0$ \KwTo $T-1$}{
                $\mathbf{u}^*_t \leftarrow \hat{\mathbf{u}}_{\tau} + \sum_{k=1}^{K}w_k \boldsymbol{\epsilon}_{k,\tau}$\;
            }
            \BlankLine
            $CommandActuators(\mathbf{u}^*_{0})$\;
            \For{$\tau \leftarrow 1$ \KwTo $T-1$}{
                $\hat{\mathbf{u}}_{\tau-1} \leftarrow\mathbf{u}^*_{\tau}$\;}
            \textcolor{drpablue}{
            \BlankLine
            \tcp{Local minimum detection}
            \uIf{using target-directed guidance cost}{
                $\mathbf{P}^* \leftarrow PredictTrajectory(\mathbf{x}_0, \mathbf{U}^*)$\;
                $\mathbf{p}_{\text{min}} \leftarrow LocalMinimumDetection\bigl(\mathbf{P}^*, \tau_{\text{monitor}}, T, r_{\text{thrs}}\bigr)$\;
                \If{$\mathbf{p}_{\textup{min}} \neq \varnothing$}{
                    $SetVirtualTarget\PP{G_{\text{detour}}, \mathbf{p}_{\text{min}}, \mathbf{p}_{\text{target}},d_{\textup{vt}}}$\;
                    $SetGuidanceCost\PP{\phi,G_{\text{detour}}}$\;
                }
            }
            \tcp{Local minimum passage detection}
            \Else{
                $\mathbf{p}_{\textrm{current}}\leftarrow ExtractPosition(\mathbf{x}_{0})$\;
                \If{$PassageDetection\PP{\mathbf{p}_{\textup{current}},\mathbf{p}_{\textup{min}},d_{\textup{vt}}}$}{
                    $SetGuidanceCost\PP{\phi,G_{\textrm{target}}}$\;
                }
            }}
        }
        \caption{\textcolor{drpablue}{DRPA}-MPPI Algotithm}
        \label{sec4:alg:drpa}
    \end{algorithm}

\subsection{Complete DRPA-MPPI Algorithm}\label{sec4_subsec:drpa-framework}
{
    Algorithm \ref{sec4:alg:drpa} presents the complete DRPA-MPPI framework.
    The algorithm initializes the target position and sets the initial guidance cost to the target-directed cost (lines 1-2). 
    The main loop (lines 3-38) continues until the robot reaches its target. 
    Within each iteration, the robot first estimates its current state and updates the obstacle map (lines 4-6).
    Lines 7-18 implement the core MPPI procedure: parallel rollout sampling and cost computation. 
    For each of the $K$ rollouts, the algorithm applies random perturbations to the control sequence, simulates the system dynamics, and computes the associated costs, including both state-dependent and control-dependent terms.
    Lines 18-24 perform importance sampling. 
    The algorithm identifies the minimum cost among all rollouts to prevent numerical issues, assigns weights to each rollout based on costs, and computes the optimal control update through a weighted average of perturbations.
    Lines 25-27 apply the updated control to the robot, while lines 38-40 shift the control sequence for the next iteration.
    This implements the receding horizon fashion of model predictive control.
    
    The core DRPA-MPPI components in lines 29-38 distinguish our approach from standard MPPI.
    During target-directed guidance (lines 29-34), the algorithm predicts the trajectory from the optimal control sequence and checks for local minima using the detection method from Section \ref{sec4_subsec:local minima detection}. 
    Upon detecting a local minimum, the algorithm switches to detour-inducing guidance by setting a virtual target and updating the cost function.
    During detour-inducing guidance (lines 35-38), the algorithm checks whether the robot has navigated past the local minimum using the passage detection from Section \ref{sec4_subsec:termination}. 
    When the passage detection condition is met, the algorithm reverts to target-directed guidance. 
    This enables the robot to resume seeking the shortest path to the target.
    This dynamic switching between guidance costs based on local minimum detection and passage detection enables DRPA-MPPI to navigate effectively around obstacles while maintaining computational efficiency.
}

%% file: src/sec5_experiments.tex
\section{Experiments}\label{sec5}

This section demonstrates the DRPA-MPPI algorithm through simulations to validate its robustness in avoiding local minima entrapment in unstructured environments.

\subsection{Experimental Setups}

\subsubsection{Task Description}

We evaluate our proposed method on a point-goal navigation task. 
The robot must navigate from a designated start position to a target position while avoiding environmental obstacles in both convex and non-convex configurations.
We consider a navigation attempt successful when the robot reaches within 0.5 m of the target position within the 30.0 s time limit.

\subsubsection{Robot Configuration}
    
We consider the kinematics model of a differential wheeled robot. 
The system transition follows discrete-time kinematics:
\begin{equation}
\begin{bmatrix}
x_{t+1}\\
y_{t+1}\\
\theta_{t+1}
\end{bmatrix}
=
\begin{bmatrix}
x_{t}\\
y_{t}\\
\theta_{t}
\end{bmatrix}
+
\begin{bmatrix}
v_{t}\cos{\theta_{t}}\\
v_{t}\sin{\theta_{t}}\\
\omega_{t}
\end{bmatrix}
\Delta t.
\end{equation}
The state vector $\mathbf{x}_t = [x_t, y_t, \theta_t]^\top$ consists of position $\mathbf{p}_{t}=[x_t, y_t]^\top$ m and heading angle $\theta_t$ rad.
The control input $\mathbf{u}_t = [v_t, \omega_t]^\top$ comprises translational velocity $v_t$ m/s and angular velocity $\omega_t$ rad/s.
Our experiments constrain $v_t$ and $\omega_t$ to $[-2.0, 2.0]$ m/s and $[-1.5, 1.5]$ rad/s, respectively.
All simulations use a fixed time step $\Delta t$ of $0.10$ s.

\subsubsection{Obstacle Configurations}

We prepare distinct obstacle configurations for both qualitative and quantitative evaluations. 
For qualitative assessment, we design three obstacle types: a short rectangular obstacle (1 m width), a long rectangular obstacle (5 m width), and a U-shaped obstacle (5 m width with 2 m indent). 
We position these obstacles symmetrically between the robot's start and target positions to create progressively challenging navigation scenarios.

For quantitative evaluation, we develop two configuration types: convex and non-convex configurations. 
We define the obstacle region as a square area ($30\,\text{m} \times 30\,\text{m}$) and partition it into either $6 \times 6$ or $10 \times 10$ grid cells in a checkered pattern.
Obstacles are placed in alternating cells to create a structured environment.
The convex configuration employs randomly generated convex polygons in each obstacle cell, created by sampling points along grid cell edges and computing their convex hull.
The non-convex configuration uses obstacles formed by combining two random convex polygons, which presents more challenging navigation conditions.
These grid configurations represent different environmental characteristics while maintaining equal obstacle density. 
The $10 \times 10$ grid creates more cluttered environments with smaller obstacles, while the $6 \times 6$ grid features fewer but larger obstacles.

To ensure statistical reliability of quantitative evaluation, we generate 1000 distinct scenarios for each obstacle type and grid resolution combination.
We randomly select the robot's starting position and the target position from their respective designated lines. 
Across all experimental trials, we initialize the robot's heading angle to point directly toward the target position to establish a consistent initial orientation.

\subsubsection{Comparable Methods}\label{sec5:sec:validation}

We evaluate our proposed DRPA-MPPI against four baseline planners: MPPI~\cite{mppi} and Log-MPPI~\cite{sampling_dist_log_mppi}, each with prediction horizons of 50 and 100. 
Log-MPPI extends standard MPPI by using NLN-Mixture instead of Gaussian distribution for sampling. 
This modification improves state space exploration with negligible computational overhead. 
This feature makes Log-MPPI an ideal baseline for DRPA-MPPI, as both methods offer performance improvements with minimal computational cost.

We use consistent parameters across all planners: $K = 10000$ sampled rollouts, temperature parameter $\lambda = 10.0$, control cost parameter $\gamma = 0.1$, and covariance matrix $\Sigma_\epsilon = \mathrm{diag}\{0.5,\,0.5\}$ for sampling distribution.
For Log-MPPI, additional lognormal distribution parameters $\boldsymbol{\mu}_{\text{ln}} = [-0.020,-0.020]$ and $\boldsymbol{\Sigma}_{\text{ln}} = \text{diag}\{0.141,0.141\}$ are selected to ensure the variance of the NLN-mixture matched $\Sigma_\epsilon$ while achieving a kurtosis of 3.25, slightly higher than that of a Gaussian distribution.
DRPA-MPPI uses specific parameters to detect and avoid local minima: monitoring window start index $\tau_{\text{monitor}} = 40$, local minima detection threshold $r_{\text{thrs}} = 0.2$ m, virtual target distance $d_{\text{vt}} = 10$ m, and termination margin $d_{\text{margin}} = 0.25$ m.

\begin{figure*}[t!]
    \centering
    \includegraphics[width=\linewidth]{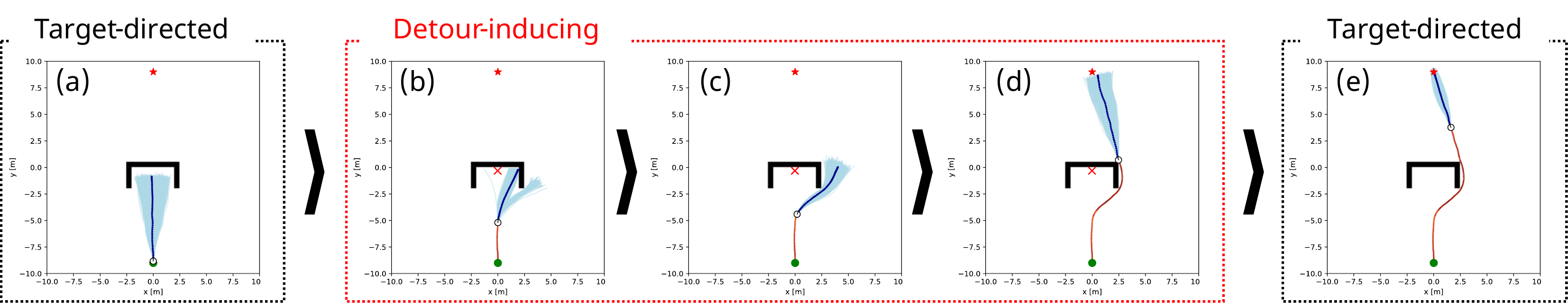}
    \caption{DRPA-MPPI in a U-shaped obstacle configuration. Green circles, red stars, and blue shading represent start and target positions, and the top 10\% of sampled rollouts, respectively. The sequence shows: (a) initial target-directed movement, (b) detection of local minimum (×) and function switch, (c) obstacle avoidance with detour function, (d) continued detour until passage criteria are met, (e) return to target-directed function after obstacle bypass.}
    \label{sec5:fig:qualitative_example_u_shaped_drpa}
\end{figure*}

\subsubsection{Evaluation Metrics}

We evaluate navigation performance using three metrics:

\begin{itemize}
    \item \textbf{Success Rate} (SR) [\%] measures navigation capability as the percentage of scenarios in which the robot reached the target position within the 30.0 s time limit.
    \item \textbf{Success Time} (ST) [s] quantifies time efficiency by averaging the time taken to reach the target across all successful navigation scenarios.
    \item \textbf{Computation Time} (CT) [ms] assesses real-time feasibility by measuring the average computation time required per control update.
\end{itemize}

\subsection{Results and Discussions}

\subsubsection{Qualitative Results}

The navigation success results for all three obstacle scenarios are summarized in Table \ref{sec5:tab:qualitative_results}.
DRPA-MPPI successfully navigates all obstacle scenarios with only a 50-step (5.0 s) horizon, including the challenging U-shaped obstacle. 
Fig. \ref{sec5:fig:qualitative_example_u_shaped_drpa} demonstrates the DRPA-MPPI navigation process in detail:
(a) The robot initially moves directly toward the target using the target-directed guidance function. 
(b) When the algorithm detects a potential local minimum inside the U-shaped cavity, it switches to the detour-inducing guidance function. This function creates repulsive forces away from the detected local minimum. 
(c) The robot then navigates around the obstacle with the detour-inducing guidance function. 
(d) The robot maintains this detour behavior until it satisfies the local minimum passage detection criterion. 
(e) After the robot bypasses the obstacle, the algorithm reverts to the target-directed guidance function for efficient movement to the target position.

In contrast, standard MPPI shows limitations in complex environments. 
With a 50-step horizon, MPPI can only navigate the short rectangular obstacle and fails with more complex geometries. 
When we extend the prediction horizon to 100 steps, MPPI becomes capable of handling the long rectangular obstacle as well. 
This demonstrates the need for extended planning horizons when MPPI confronts larger obstacles. 
Log-MPPI produces identical results to MPPI and is therefore omitted from the table.

\begin{table}[t]
\centering
\caption{Navigation success (\checkmark) across obstacle scenarios}
\label{sec5:tab:qualitative_results}
\begin{tabular}{l|ccc}
\toprule
Planner Name & Short Rect. & Long Rect. & U-shaped \\
\midrule
MPPI (horizon: 50) & \checkmark & & \\
MPPI (horizon: 100) & \checkmark & \checkmark & \\
\textbf{DRPA-MPPI (horizon: 50)} & \textbf{\checkmark} & \textbf{\checkmark} & \textbf{\checkmark} \\
\bottomrule
\end{tabular}
\end{table}

\begin{table*}[t]
\centering
\begin{threeparttable}        
\caption{Quantitative results for both random convex and non-convex obstacle scenarios}
\label{sec5:tab:quantitative_results}
\begin{tabular}{@{}l c *{12}{c}@{}}
\toprule
& & \multicolumn{6}{c}{Random convex obstacle} & \multicolumn{6}{c}{Random non-convex obstacle} \\
\cmidrule(lr){3-8} \cmidrule(l){9-14}
& & \multicolumn{3}{c}{$10 \times 10$ grid} & \multicolumn{3}{c}{$6 \times 6$ grid} & \multicolumn{3}{c}{$10 \times 10$ grid} & \multicolumn{3}{c}{$6 \times 6$ grid} \\
\cmidrule(lr){3-5} \cmidrule(lr){6-8} \cmidrule(lr){9-11} \cmidrule(l){12-14}
Planner & Horizon & SR $\uparrow$ & ST $\downarrow$ & CT $\downarrow$ & SR $\uparrow$ & ST $\downarrow$ & CT $\downarrow$ & SR $\uparrow$ & ST $\downarrow$ & CT $\downarrow$ & SR $\uparrow$ & ST $\downarrow$ & CT $\downarrow$ \\
\midrule
MPPI~\cite{mppi} & 50 & 94.2 & 22.8 & \textbf{16.5} & 81.7 & 23.6 & \textbf{16.6} & 80.3 & 24.2 & \textbf{16.6} & 61.8 & 25.3 & \textbf{16.4} \\
 & 100 & 91.2 & 27.3 & 32.7 & 92.0 & 26.6 & 32.7 & 76.2 & 28.3 & 32.2 & 84.3 & 27.3 & 32.6 \\
\midrule
Log-MPPI~\cite{sampling_dist_log_mppi} & 50 & 94.3 & 22.8 & 17.8 & 81.8 & 23.7 & 18.0 & 81.0 & 24.2 & 17.8 & 60.9 & 25.4 & 18.1 \\
 & 100 & 90.9 & 27.2 & 35.8 & 91.8 & 26.6 & 35.2 & 74.4 & 28.3 & 35.7 & 84.0 & 27.3 & 35.8 \\
\midrule
\rowcolor[gray]{0.9}
\textbf{Ours: DRPA-MPPI} & \textbf{50} & \textbf{99.7} & \textbf{22.6} & 17.6 & \textbf{99.9} & \textbf{22.4} & 17.9 & \textbf{98.0} & \textbf{23.1} & 17.8 & \textbf{97.3} & \textbf{22.9} & 17.3 \\
\bottomrule
\end{tabular}
\begin{tablenotes}[flushleft]
\item SR = Success Rate (\%), ST = Success Time (s), CT = Computation Time (ms). Results based on 1000 distinct scenarios for each obstacle configuration and grid resolution combination. 
\end{tablenotes}
\end{threeparttable}
\end{table*}

\begin{figure*}[t]
    \centering
    \includegraphics[width=\linewidth]{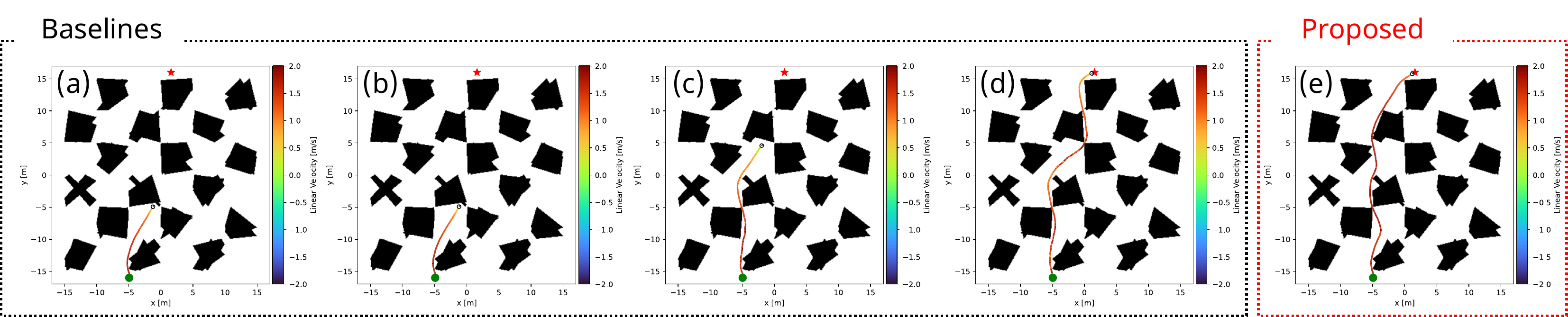}
    \caption{Example trajectories in a $6 \times 6$ grid non-convex obstacle scenario. Green dots, red stars, and color gradients respectively show start and target positions, and linear velocity magnitude. Results: (a) MPPI (horizon: 50) and (b) Log-MPPI (horizon: 50) both trapped at same local minima; (c) MPPI (horizon: 100) escapes the initial local minimum but encounters another trap, failing to reach the target; (d) Log-MPPI (horizon: 100) reaches the target but with inefficient navigation and velocity degradation; and (e) DRPA-MPPI (horizon: 50) navigates efficiently to the target without velocity degradation.}
    \label{sec5:fig:quantitative_example_non_convex_6_6}
\end{figure*}

\subsubsection{Quantitative Results}

The quantitative results summarized in Table \ref{sec5:tab:quantitative_results} demonstrate that DRPA-MPPI consistently outperforms all baseline planners in terms of success rate (SR), with near-perfect performance ($>$ 97\%) across all test configurations regardless of obstacle size or environment clutter. 
This contrasts with the variable performance of other methods across different environmental configurations. 
Although DRPA-MPPI does not achieve 100\% success rate in convex obstacle scenarios as expected from Section \ref{sec4_subsec:detour-inducing function}, analysis of the failed scenarios reveals that each failure occurs due to incidental non-convexities formed by overlapping vertices of adjacent convex obstacles.
Fig. \ref{sec5:fig:quantitative_example_non_convex_6_6} provides visual evidence of the navigation performance differences in a representative $6 \times 6$ grid non-convex obstacle scenario. 
The trajectories illustrate the local minima entrapment issues faced by the baseline methods and demonstrate how DRPA-MPPI successfully navigates to the target with efficient velocity management despite using only a 50-step horizon. 
This visual evidence confirms DRPA-MPPI's superior capability in handling complex obstacle scenarios.

DRPA-MPPI also demonstrates superior time efficiency with the lowest success times (ST) across all configurations, as shown in Table \ref{sec5:tab:quantitative_results}. 
This result is significant given that the dynamic repulsive potential augmentation promotes detour behaviors.
The improved time efficiency can be attributed to two key factors: 1) rapid and accurate local minima detection that enables immediate escape from potential entrapment situations, and 2) the effective termination condition that transitions back to target-directed navigation once the local minimum is successfully bypassed.

Regarding computational efficiency, Table \ref{sec5:tab:quantitative_results} shows that DRPA-MPPI achieves computation times (CT) comparable to MPPI and Log-MPPI with equivalent 50-step horizons (17-18 ms), while significantly outperforming longer-horizon variants (32-36 ms). 
These results confirm that DRPA-MPPI improves obstacle avoidance without introducing notable computational overhead compared to standard methods with equivalent horizons, while substantially reducing computational requirements compared to longer-horizon methods.

The baseline planners exhibit performance variations strongly correlated with environmental characteristics, as summarized in Table \ref{sec5:tab:quantitative_results}. 
Shorter-horizon planners (MPPI and Log-MPPI with 50-step horizons) show higher success rates in the $10 \times 10$ grid configurations compared to the $6 \times 6$ grid configurations, which indicates these planners struggle more with larger obstacles than with more cluttered environments. 
Conversely, longer-horizon planners (MPPI and Log-MPPI with 100-step horizons) perform better in $6 \times 6$ grid configurations than in $10 \times 10$ grid configurations. 
This suggests that extended prediction horizons enable better navigation around larger obstacles but offer less advantage in more cluttered environments with numerous smaller obstacles.

The quantitative results and qualitative observations demonstrate that DRPA-MPPI achieves better navigation robustness while maintaining computational efficiency. 
This makes it particularly well-suited for autonomous navigation in unstructured environments with complex obstacle scenarios of varying sizes and distributions.

%% file: src/sec6_conclusion.tex
\section{Conclusion}
{
    This paper presented DRPA-MPPI, a framework that addresses local minima issues well-known in standard MPPI-based navigation.
    The proposed framework achieves local minima-free robot navigation in unstructured and unknown environment by automatically detecting and avoiding potential entrapment situations without sacrificing computational efficiency.
    Our approach identifies entrapment situations through trajectory prediction analysis to trigger transitions between target-directed and detour-inducing guidance functions by augmenting repulsive potential terms to the cost function.
    Extensive simulations in both convex and non-convex obstacle environments show that DRPA-MPPI outperforms standard MPPI and its variants in navigation success rate and time efficiency while maintaining comparable computational performance.
    Our theoretical analysis further guarantees the method's capability to escape from local minima in convex obstacle scenarios.

    Future work will extend DRPA-MPPI in three key directions.
    First, real-world validation is needed to verify the method's robustness under actual sensing and actuation constraints. 
    Second, theoretical guarantees should be expanded to non-convex obstacles. 
    Finally, extending implementation to other platforms with complex three-dimensional dynamics would broaden its applicability.
    Integration with global replanning strategies will also advance autonomous navigation in highly unstructured environments.
}

%% file: src/appendix.tex
\appendix
\section{Appendix}\label{appendix}
{
    This appendix provides proofs for the properties of $G_{\text{detour}}$ discussed in Sec. \ref{sec4_subsec:detour-inducing function}. 
    As a preliminary step, we present the partial derivatives of $G_{\text{detour}}$ for $\mathbf{p} \not\in \{\mathbf{p}_{\text{min}}, \mathbf{p}_{\text{vt}}\}$:
    \begin{equation}
        \nabla G_{\text{detour}}({\bf{p}}) = \frac{\mathbf{p}-\mathbf{p}_{\text{vt}}}{\|\mathbf{p}_{\text{vt}}-\mathbf{p}\|}-w_{\text{rep}}\cdot\frac{\mathbf{p}-\mathbf{p}_{\text{min}}}{\|\mathbf{p}_{\text{min}}-\mathbf{p}\|}
    \end{equation}
    \begin{property1}
        \(\mathbf{p}_{\text{vt}}\) is both the unique local and global minimum when \(0 < w_{\text{rep}} < 1\).
    \end{property1}
    \begin{proof}
    {
        First, consider $\mathbf{p} \in \{\mathbf{p}_{\text{min}}, \mathbf{p}_{\text{vt}}\}$ where \(G_{\text{detour}}\) is not differentiable. 
        Since $w_{\text{rep}} > 0$, we have $G_{\text{detour}}(\mathbf{p}_{\text{vt}}) < G_{\text{detour}}(\mathbf{p}_{\text{min}})$.

        Next, for $\mathbf{p}\not\in\{\mathbf{p}_{\text{min}}, \mathbf{p}_{\text{vt}}\}$ where \(G_{\text{detour}}\) is differentiable, we seek conditions under which ${-\nabla G_{\text{detour}}({\bf{p}})\cdot(\mathbf{p}_{\text{vt}}-\mathbf{p})>0}$.
        This condition indicates that \(G_{\text{detour}}\) decreases in the direction toward $\mathbf{p}_{\text{vt}}$. 
        Through algebraic manipulation, we obtain:
        \begin{align}
            &{-\nabla G_{\text{detour}}({\bf{p}})\cdot(\mathbf{p}_{\text{vt}}-\mathbf{p})>0}  \nonumber \\
            &\Leftrightarrow \|\mathbf{p}_{\text{vt}} - \mathbf{p}\| - w_{\text{rep}}\frac{(\mathbf{p}_{\text{min}} - \mathbf{p}) \cdot (\mathbf{p}_{\text{vt}} - \mathbf{p})}{\|\mathbf{p}_{\text{min}} - \mathbf{p}\|} > 0 \nonumber \\
            &\Leftrightarrow \|\mathbf{p}_{\text{vt}} - \mathbf{p}\|(1 - w_{\text{rep}}\cos\theta)>0 \nonumber \\
            &\Leftrightarrow 1 - w_{\text{rep}}\cos\theta>0. \label{appendix:eq:inner_prop1}
        \end{align}
        Here, $\theta$ is the angle between $\mathbf{p}_{\text{vt}} - \mathbf{p}$ and $\mathbf{p}_{\text{min}} - \mathbf{p}$, with $0\leq\theta\leq\pi$. 
        Since $1-w_{\text{rep}} \leq 1 - w_{\text{rep}}\cos\theta \leq 1+w_{\text{rep}}$, the condition in \eqref{appendix:eq:inner_prop1} is always satisfied when $w_{\text{rep}}<1$.

        In summary, when $0<w_{\text{rep}}<1$, for all differentiable we have ${-\nabla G_{\text{detour}}({\bf{p}})\cdot(\mathbf{p}_{\text{vt}}-\mathbf{p})>0}$, and at non-differentiable points, $G_{\text{detour}}(\mathbf{p}_{\text{vt}}) < G_{\text{detour}}(\mathbf{p}_{\text{min}})$. 
        Thus, \(\mathbf{p}_{\text{vt}}\) is the unique local and global minimum under these conditions.
    }
    \end{proof}
    \begin{property2}
        Suppose $\mathbf{p}_{\text{n}}$ satisfies $(\mathbf{p}_{\textup{vt}} - \mathbf{p}_{\textup{min}}) \cdot (\mathbf{p}_{\text{n}} - \mathbf{p}_{\textup{min}}) = 0$ and $\mathbf{p}_{\text{n}} \neq \mathbf{p}_{\text{min}}$. 
        Then $-\nabla G_{\textup{detour}}(\mathbf{p}_{\text{n}}) \cdot (\mathbf{p}_{\text{n}} - \mathbf{p}_{\textup{min}}) \geq 0$ holds if $\|\mathbf{p}_{\textup{min}} - \mathbf{p}_{\text{n}}\| \leq d_{\textup{virtual}} \PP{\frac{w_{\textup{rep}}^2}{1 - w_{\textup{rep}}^2}}^{\frac{1}{2}}$.
    \end{property2}

    \begin{proof}
        We manipulate $-\nabla G_{\textup{detour}}(\mathbf{p}_{\text{n}}) \cdot (\mathbf{p}_{\text{n}} - \mathbf{p}_{\textup{min}}) \geq 0$ as follows:
        \begin{align}
            &-\nabla G_{\textup{detour}}(\mathbf{p}_{\text{n}}) \cdot (\mathbf{p}_{\text{n}} - \mathbf{p}_{\textup{min}}) \geq 0 \nonumber \\
            &\Leftrightarrow 
            \frac{(\mathbf{p}_{\text{vt}} - \mathbf{p}_{\text{n}}) \cdot (\mathbf{p}_{\text{n}} - \mathbf{p}_{\textup{min}})}{\|\mathbf{p}_{\text{vt}} - \mathbf{p}_{\text{n}}\|} 
            - w_{\text{rep}}\|\mathbf{p}_{\text{n}} - \mathbf{p}_{\textup{min}}\| 
            \geq 0 \nonumber \\
            &\Leftrightarrow 
            \frac{\|\mathbf{p}_{\text{n}} - \mathbf{p}_{\textup{min}}\|^2 }{\|(\mathbf{p}_{\text{vt}}-\mathbf{p}_{\text{min}}) - (\mathbf{p}_{\text{n}}-\mathbf{p}_{\text{min}})\|} 
            - w_{\text{rep}}\|\mathbf{p}_{\text{n}} - \mathbf{p}_{\textup{min}}\| 
            \geq 0 \nonumber \\
            &\Leftrightarrow 
            \frac{\|\mathbf{p}_{\text{n}} - \mathbf{p}_{\textup{min}}\|^2}{\|\mathbf{p}_{\text{n}} - \mathbf{p}_{\textup{min}}\|^2+d_{\text{vt}}^2} 
            - w_{\text{rep}}\|\mathbf{p}_{\text{n}} - \mathbf{p}_{\textup{min}}\| 
            \geq 0 \nonumber \\
            &\Leftrightarrow 
            \|\mathbf{p}_{\textup{min}} - \mathbf{p}_{\text{n}}\| \leq d_{\textup{virtual}} \PP{\frac{w_{\textup{rep}}^2}{1 - w_{\textup{rep}}^2}}^{\frac{1}{2}}.
            \label{appendix:eq:inner_prop2}
        \end{align}

        Therefore, for any $\mathbf{p}_{\text{n}}$ satisfying $(\mathbf{p}_{\textup{vt}} - \mathbf{p}_{\textup{min}}) \cdot (\mathbf{p}_{\text{n}} - \mathbf{p}_{\textup{min}}) = 0$ and $\mathbf{p}_{\text{n}} \neq \mathbf{p}_{\text{min}}$, we have proven that $-\nabla G_{\textup{detour}}(\mathbf{p}_{\text{n}}) \cdot (\mathbf{p}_{\text{n}} - \mathbf{p}_{\textup{min}}) \geq 0$ if and only if $\|\mathbf{p}_{\textup{min}} - \mathbf{p}_{\text{n}}\| \leq d_{\textup{virtual}} \PP{\frac{w_{\textup{rep}}^2}{1 - w_{\textup{rep}}^2}}^{\frac{1}{2}}$.
    \end{proof}